\documentclass{bmvc2k}

\pdfoutput=1
\title{End-to-End Lane detection with One-to-Several Transformer}

\addauthor{Kunyang Zhou}{kunyangzhou@seu.edu.cn}{1}
\addauthor{Rui Zhou}{2130110210@stmail.ntu.edu.cn}{2}

\addinstitution{
 Southeast University\\
 Nanjing, China
}
\addinstitution{
 Nantong University\\
 Nantong, China
}
\runninghead{Zhou,Zhou}{End to End Lane detection with One-to-Several Transformer}

\begin{document}

\maketitle

\vspace{-0.7cm}

\begin{abstract}
   Although lane detection methods have shown impressive performance in real-world scenarios, 
   most of methods require post-processing which is not robust enough. 
   Therefore, end-to-end detectors like DEtection TRansformer(DETR) have been introduced in lane detection. 
   However, one-to-one label assignment in DETR can degrade the training efficiency due to label semantic conflicts.   
   Besides, positional query in DETR is unable to provide explicit positional prior, 
   making it difficult to be optimized. In this paper, we present the One-to-Several Transformer(O2SFormer)\footnote{\noindent Code: \href{https://github.com/zkyseu/O2SFormer}{https://github.com/zkyseu/O2SFormer}}. 
   We first propose the one-to-several label assignment, which combines one-to-many and one-to-one label assignment to 
   solve label semantic conflicts while keeping end-to-end detection. To overcome the difficulty in 
   optimizing one-to-one assignment. We further propose the layer-wise soft label 
   which dynamically adjusts the positive weight of positive lane anchors in different decoder layers. 
   Finally, we design the dynamic anchor-based positional query to explore positional prior 
   by incorporating lane anchors into positional query. Experimental results show that O2SFormer with ResNet50 backbone achieves 77.83\% F1 score on CULane dataset, 
   outperforming existing Transformer-based and CNN-based detectors. Futhermore, O2SFormer converges 12.5× faster than DETR for the ResNet18 backbone. 
\vspace{-0.44cm}
\end{abstract}

\begin{figure*}[!h]
   \begin{center}
   \includegraphics[height = 3cm,width=11cm]{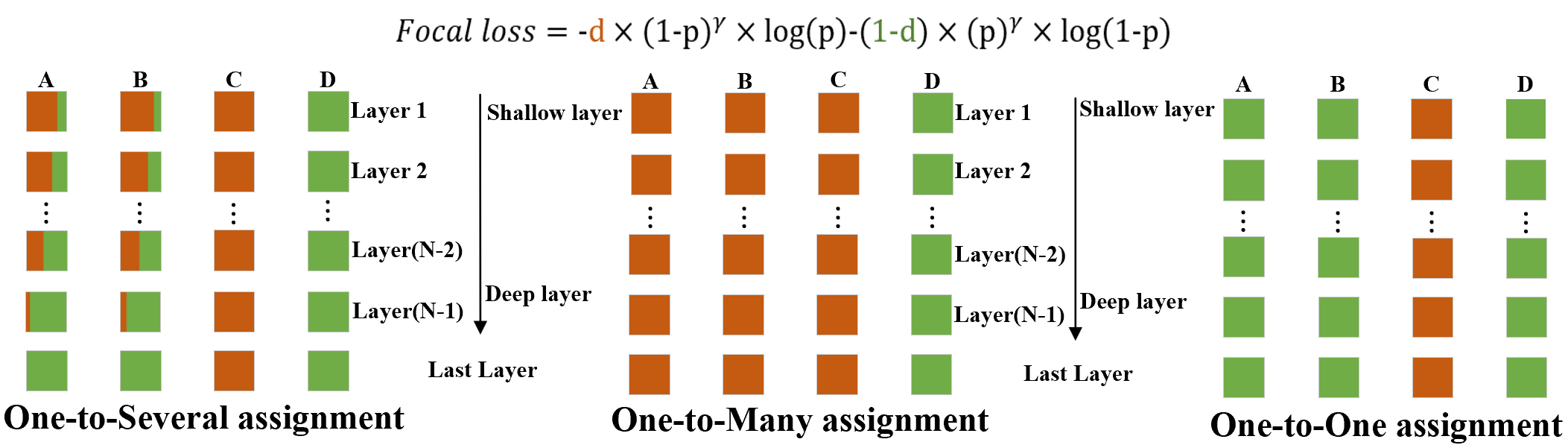}
   \vspace{-0.5cm} 
   \end{center}
      \caption{The positive and negative weights of different lane anchors(A,B,C and D) in the classification 
      loss(such as focal loss~\cite{lin2017focal}) across different decoder layers. Each lane anchor has a positive weight d(in brown color) and 
      a negative weight 1-d (in green color). C is a fully positive lane anchor, D is a fully negative lane anchor. 
      A and B have the similar label semantic as C. In one-to-one(o2o) assignment, A and B are assigned as negative lane anchor, 
      which leads to label semantic conflicts. While in one-to-many(o2m) assignment, A and B are assigned as positive lane anchor, 
      which requires post-processing to remove duplicate predictions. Moreover, in o2o and o2m assignment, the positive weights for all lane anchors are fixed.
      In our one-to-several assignment, the positive weights of A and B are dynamically adjusted according to the decoder layers.}
   \vspace{-0.4cm}
   \label{fig:short1}
   \end{figure*}

   \section{Introduction}
   \label{sec:intro}
   \vspace{-0.2cm}
   Lane detection is a fundamental yet challenging task in computer vision~\cite{hou2019learning,wang2000lane,narote2018review}, 
   which plays an important role in many applications, such as lane keeping, 
   adaptive cruise control and driving route planning. Recently, many studies~\cite{lee2022robust,xu2020curvelane,chen2019pointlanenet,zheng2021resa} 
   focus on using convolutional neural networks (CNNs) for lane detection, 
   which are significantly superior to traditional methods~\cite{niu2016robust,jung2013efficient,borkar2011novel} in performance. 
   Nevertheless, most of the state-of-the-art lane detection methods~\cite{zheng2022clrnet,li2019line,tabelini2021keep} depend 
   on post-processing operations, such as non-maximum suppression (NMS). 
   NMS is not robust enough and difficult to optimize~\cite{lv2023detrs}. In addition, due to the limitation of receptive field, 
   CNNs have no advantage in capturing long-range dependence, leading to inefficient feature learning.
   
   DETR~\cite{carion2020end} is an end-to-end object detector based on Transformer, which treats object detection as a set prediction problem. 
   The self-attention module in Transformer enables DETR to capture long-range dependence better than CNNs. 
   Although there have been several attempts~\cite{han2022laneformer,liu2021end} to employ DETR in lane detection, existing methods need either 
   the camera parameters~\cite{liu2021end} or an additional object detector~\cite{han2022laneformer}, which hinders the generalization and flexibility 
   of DETR. On the other hand, directly applying DETR to lane detection results in slow convergence and significant 
   performance gaps compared to CNNs. We attribute reasons to two main factors: (1) the positional query in 
   DETR lacks a clear focus area~\cite{meng2021conditional}, making it difficult for the decoder to locate thin and elongated lanes, 
   and (2) the one-to-one label assignment causes low training efficiency due to label semantic conflicts(as seen in Fig.~\ref{fig:short1}).
   
   In this paper, we propose One-to-Several Transformer(O2SFormer) to address the aforementioned issues. 
   Firstly, to explore explicit positional prior for the positional query, we design the dynamic anchor-based 
   positional query, where lane anchors are encoded as positional query. We update lane anchors layer-by-layer 
   to improve the localization accuracy of lane anchors. Next, we introduce the One-to-Several (O2S) 
   label assignment to solve label semantic conflicts, which combines one-to-many and one-to-one assignment. 
   As illustrated in Fig.~\ref{fig:short1}, we utilize one-to-many assignment for the first N-1 decoder layers and one-to-one assignment 
   for the last layer. Due to the varying learning capabilities of different decoder layers, assigning the same 
   positive weight to all positive lane anchors in the first N-1 decoder layers would make it difficult to optimize 
   the one-to-one assignment. Therefore, we further propose the layer-wise soft label. We assign soft labels with high positive 
   weights to positive lane anchors in the shallow layers of the decoder to improve the discrimination of feature representation, 
   while assigning soft labels with low positive weights to positive lane anchors in the deep layers of the decoder to emphasize end-to-end detection. 
   By adopting the layer-wise soft label, we achieve a better balance between one-to-many and one-to-one assignment. 
   
   In summary, the contributions of this paper are summarized as follows.
   \vspace{-0.12cm}
   \begin{itemize}
      \item We design a novel one-to-several label assignment which combines the advantages of one-to-one and one-to-many assignment
      to alleviate label semantic conflicts while keeping end-to-end detection. Moreover, we propose the layer-wise soft label to balance one-to-one and one-to-many assignment.
      \vspace{-0.12cm}
      \item We present a novel dynamic anchor-based positional query to explore explicit positional prior, which encodes lane anchors as positional queries and updates lane anchors layer-by-layer.
      \vspace{-0.12cm}
      \item We evaluate the O2SFormer on the CULane dataset. Experimental results show that O2SFormer accelerates DETR training and achieves better performance than Transformer-based and CNN-based detectors.
      \end{itemize}

\section{Related Work}
\label{sec:Related Work}
\vspace{-0.2cm}
\subsection{Lane detection}
Deep learning-based lane detection can be divided into three categories according to the representation of lane: 
segmentation-based method, anchor-based method, and parameter-based method. Segmentation-based methods~\cite{pan2018spatial,zheng2021resa,yang2023lane} 
perform pixel-wise prediction such as SCNN~\cite{pan2018spatial}, while segmentation-based methods ignore to take the lane 
as a whole, resulting in insuperior performance. Anchor-based methods~\cite{li2019line,tabelini2021keep,zheng2022clrnet} regress accurate lanes 
by refining predefined anchors. Different from segmentation-based and anchor-based methods, 
parameter-based methods~\cite{tabelini2021polylanenet,liu2021end} consider lane as a polynomial function and perform lane 
detection by regressing the parameters of the polynomial function. Since anchor-based methods 
are more robust and accurate than the other two methods, 
we implement O2SFormer based on the anchor-based method.
\vspace{-0.31cm}
\subsection{Label assignment in DETR}
DETR~\cite{carion2020end} performs one-to-one label assignment to achieve end-to-end object detection. 
Recent works~\cite{chen2022group,jia2022detrs,zhang2023dense} have shown that one-to-one label assignment produces low training efficiency because of sparse supervision and label semantic conflicts.
Group DETR~\cite{chen2022group} introduces additional queries to increase 
supervision. H-DETR~\cite{jia2022detrs} is similar to Group DETR, except that the additional queries are 
assigned labels by one-to-many assignment. However, introducing additional queries brings out 
extra computational burden during training. O2SFormer employs one-to-many assignment in the first N-1 decoder layers to increase supervision without introducing extra queries.
DDQ~\cite{zhang2023dense} filters out anchors with similar label semantic as fully positive anchors by a simple class-agnostic NMS. Unlike DDQ, 
O2SFormer assigns layer-wise soft labels to positive lane anchors with similar label semantic as fully positive lane anchors.
\vspace{-0.2cm}

\section{Method}
\begin{figure*}[!h]
   \begin{center}
   \includegraphics[height = 5.5cm,width=13cm]{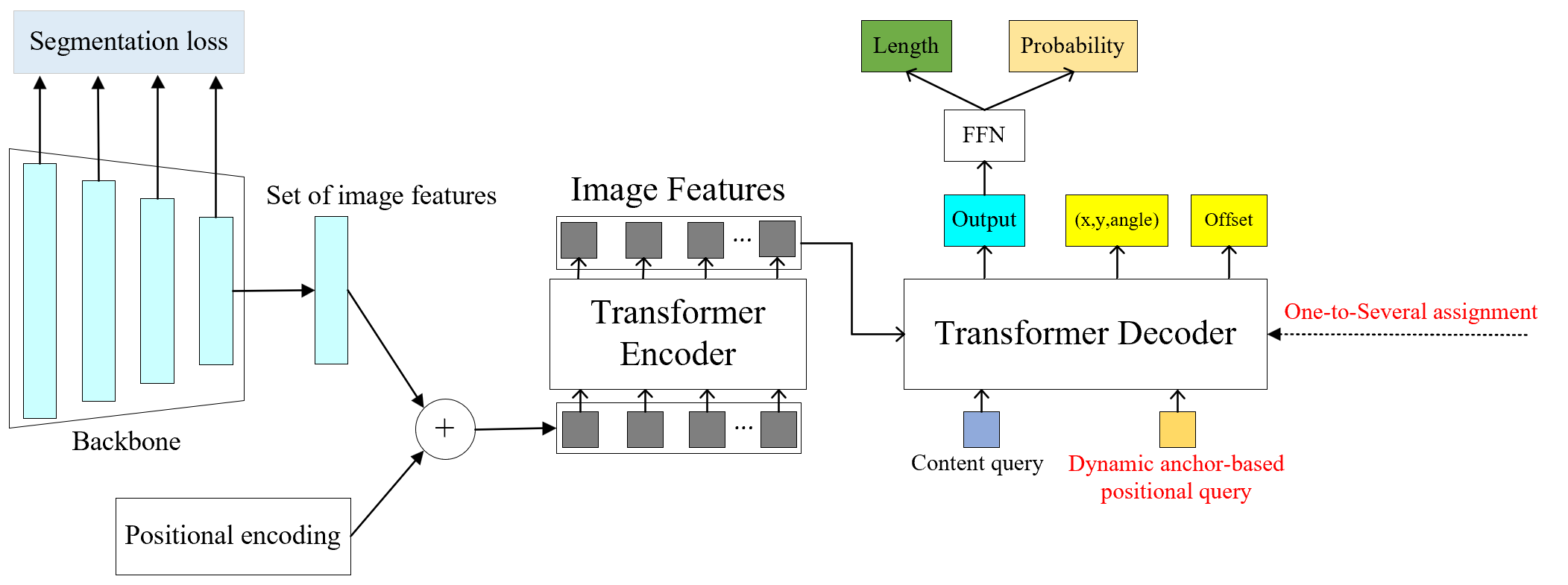}
   \vspace{-0.5cm} 
   \end{center}
      \caption{Pipeline of O2SFormer. O2SFormer adopts a convolutional neural network as the backbone 
               to learn the 2D feature representation of the input image. We flatten the 2D feature 
               representation and add positional encoding. Then, the features are fed into the Transformer 
               Encoder. The Transformer Decoder takes the dynamic anchor-based positional query, content query 
               and image features as inputs. We pass output embedding of the Transformer decoder to a feed forward network(FFN) 
               that predicts background and foreground probabilities and the length of lane anchors. One-to-Several assignment is used to assign labels for lane anchors in each decoder layer.}
   \vspace{0.2cm}
   \label{fig:short2}
   \end{figure*}

\vspace{-0.2cm}
\label{sec:method}
The structure of O2SFormer is shown in Fig.~\ref{fig:short2}. Sec~\ref{sec:rep} introduces the representation of lane anchors. 
Sec~\ref{sec:dynamic} details the Dynamic anchor-based positional query. Sec~\ref{sec:assignment} describes the One-to-several label assignment and layer-wise soft label.
\vspace{-0.4cm}
\subsection{Representation of lane anchors}
\label{sec:rep}
Lanes have obvious shape and position priors, thus predefining lane anchors on the input 
image can help the model better locate the lanes. Following~\cite{zheng2022clrnet}, we employ equally-spaced 
2D points as the representation for lane anchors and all lane anchors are learnable. 
Specifically, lane anchors can be represented as a sequence of 2D points, i.e., $ Anchor=\{(x_{1},y_{1}),...,(x_{z},y_{z})\} $.
The y-coordinate is uniformly sampled along the vertical edge of the image, i.e.,
$y_{i}=\frac{H}{z-1}*i $. Where $z$ is the number of sampling points and $H$ denotes height of the image.
The x-coordinates are one-to-one corresponding to the y-coordinates. In this paper, 
we regress the accurate lanes by refining lane anchors. The outputs of the network 
consist of four components: (1) the length of the lane anchors. (2) the start point of 
the lane anchors and the angle between x-axis of the lane anchors(denoted as $x,y$ and $\theta$). 
(3) probabilities of background and foreground. (4) The $z$ offsets, i.e., the horizontal distance 
between $z$ sampling points and its ground truth. We first predict the background and foreground 
probabilities for lane anchors. Then, we obtain the start point, length and angle 
for foreground lane anchors. Finally, we refine foreground lane anchors by regressing $z$ offsets.

\subsection{Dynamic anchor-based positional Query}
\label{sec:dynamic}
Conditional DETR~\cite{meng2021conditional} points that object query in DETR can be divided into content 
query and positional query. Content query is responsible for learning the content 
information of the image and positional query is used to learn location information. 
Conditional DETR shows that the positional query only provides the general attention 
map without giving explicit positional prior information, which makes 
network require more training epochs to learn how to locate objects. Compared with object detection, 
lanes have stronger positional prior. General attention makes network more difficult to learn the feature of lanes. 

\begin{figure*}[!h]
   \begin{center}
   \includegraphics[height = 7.5cm,width=0.47\textwidth]{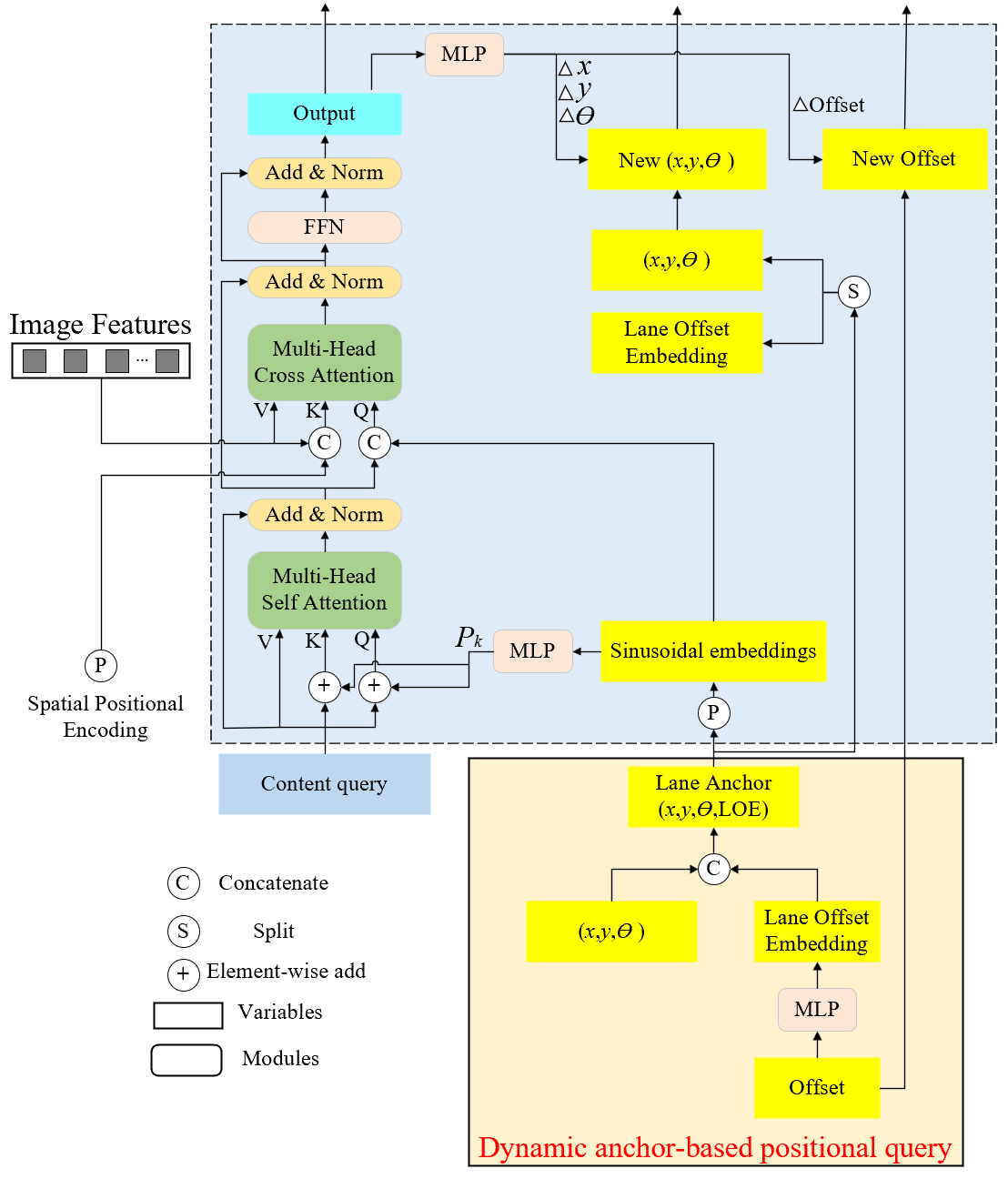}
   \vspace{-0.5cm} 
   \end{center}
      \caption{Structure of Decoder Layer. Q, K and V represent Query, Key and Value.}
      \vspace{-0.5cm}
   \label{fig:short3}
   \end{figure*}

As illustrated in Fig.~\ref{fig:short3}, dynamic anchor-based positional query aims to explore clear positional prior information. 
Since the length of the lane is not the main determinants of positional information, 
we consider the start point, angle and offset as the positional prior information of the lane.
Specifically, we use $A_k=\left(x_k,y_k,\theta_k\right)$ to represent the start point coordinate and angle of the k-th lane anchor. $offset_k$ indicates $z$ offsets of the k-th lane anchor.
Dynamic anchor-based positional query is denoted as $ P_k\in R^D$ and $D$ is dimension. 

In order to reduce the computational burden in the subsequent spatial positional encoding,
we map $offset_k$ to a float through a multilayer perceptron (MLP).
The mapped float is named with Lane Offset Embedding(${LOE}_k\in R$). We concatenate the ${LOE}_k$ with $A_k$ to 
obtain the k-th lane anchor(denoted as $M_k$).

\vspace{-0.35cm}

\begin{equation}\label{eqn-1} 
   M_k=Cat\left(A_k,{LOE}_k\right)
 \end{equation}
Where cat is concatenate operation. $P_k$ can be generated by
\begin{equation}\label{eqn-2} 
   P_k=MLP\left(PE\left(M_k\right)\right)  
 \end{equation}
PE represents spatial positional encoding to generate sinusoidal embeddings from $M_k$. As $M_k$ is a quaternion, PE is overloaded here: 
\begin{equation}\label{eqn-3} 
   PE\left(M_k\right)=PE\left(x_k,y_k,\theta_k,{LOE}_k\right)=Cat\left(PE\left(x_k\right),PE\left(y_k\right),PE\left(\theta_k\right),PE\left({LOE}_k\right)\right) 
 \end{equation}
 In spatial positional encoding, we map a float to a vector with D/2 dimensions as: ${R\rightarrow R}^\frac{D}{2}$. 
MLP maps the vector generated by PE into D dimensions. All MLP modules in the decoder layer are composed of a linear layer and ReLU activation. 

In multi-head self-attention, the content query(denoted as $C_k$) of the keys, queries, and values are the same. Queries and keys include extra position items.
\begin{equation}\label{eqn-4} 
   {Key}_k=C_k+P_k,\ \ {Query}_k=C_k+P_k,\ {Value}_k=C_k 
 \end{equation}

For multi-head cross-attention, we concatenate the content and position information as queries and keys. We use F to represent image features.
\begin{equation}\label{eqn-5} 
   {Query}_k=Cat\left(C_k,PE\left(M_k\right)\right),\ {Key}=Cat\left(F,PE\left(F\right)\right),\ {Value}=F
 \end{equation}

{\bf Dynamic update of lane anchor}. Since the initial lane anchors are not accurate enough, we update lane anchors layer-by-layer. 
In particular, as shown in Fig.~\ref{fig:short3}, we use an extra MLP module to predict $\left ( \Delta x,\Delta y, \Delta \theta,\Delta offset \right )$ from the output of the decoder layer. 
Then we split $M_k$ to get the $A_k$. The new $A_k$ and $offset_k$ are obtained as following:
\begin{equation}\label{eqn-6} 
   A_k^{new}=\ A_k+\left ( \Delta x,\Delta y, \Delta \theta \right ) 
   \vspace{-0.39cm}
 \end{equation}
\begin{equation}\label{eqn-7} 
   offset_k^{new}=offset_k + \Delta offset
 \end{equation}

\subsection{One-to-Several label assignment}
\label{sec:assignment}
DETR employs one-to-one assignment to achieve end-to-end detection. In one-to-one 
assignment, a ground truth is assigned only one positive lane anchor, while many lane anchors 
with similar label semantic as positive lane anchor are assigned as negative lane anchors, resulting in label semantic  
conflicts. Although one-to-many assignment can effectively solve label semantic conflicts, 
one-to-many assignment is unable to keep end-to-end detection. 
Hence, we propose the One-to-Several (O2S) label assignment to alleviate label semantic conflicts while keeping end-to-end detection. 

As shown in Fig.~\ref{fig:short1}, we apply one-to-many assignment in the first N-1 decoder layers and use one-to-one assignment in the last decoder layer. Existing methods~\cite{zhang2020bridging,kim2020probabilistic} assign labels separately for each decoder layer, 
leading to inconsistent label assignment across different decoder layers. On the contrary, O2S assignment
utilizes the output of the last decoder layer to assign labels for all decoder layers. 
Specifically, we first perform Optimal Transport Assignment(OTA)~\cite{ge2021ota} on the output of the last decoder layer to get the positive lane anchors for one-to-many assignment. 
OTA dynamically assigns $t\left(t\geq 2\right)$ positive lane anchors for each label
according to the Line-IOU(LIOU)~\cite{zheng2022clrnet} between the predictions and the ground truths.  
Then, we perform Hungarian matching~\cite{carion2020end} on $t$ positive lane anchors to obtain the positive lane anchor for one-to-one assignment.
We define the positive lane anchor used for one-to-one assignment as the fully positive lane anchor. 

{\bf Layer-wise soft label}. Although the first N-1 decoder layers employ one-to-many assignment to 
address label semantic conflicts, such manner makes one-to-one assignment difficult to optimize. 
To elegantly combine the advantages of one-to-one and one-to-many assignment, we propose the 
layer-wise soft label. As illustrated in Fig.~\ref{fig:short1}, we decrease the positive weights of positive lane anchors used for one-to-many assignment
layer-by-layer, except the fully positive lane anchor. For the shallow layers of decoder, 
we assign soft label with high positive weights to positive lane anchors to enhance 
the ability of feature representation. In the deep layers of decoder, we assign soft label with low 
positive weights to positive lane anchors to make the decoder focus on one-to-one assignment. For the $r$-th($1 \leq r \leq N-1$) decoder layer, the soft label of the $j$-th positive lane anchor is
\vspace{-0.23cm}
\begin{equation}\label{eqn-8} 
   \vspace{-0.39cm}
   d_j^r=\begin{cases}
                1, \text{if fully positive lane anchor} \\
               \ \frac{N-r}{N-1}\times W_j, \text{otherwise} \\
            \end{cases}
 \end{equation}
\vspace{0.05cm}
\begin{equation}\label{eqn-9} 
   W_j=\frac{{pred}_j^r}{max\left({pred}_i^r\right)},i\in S
 \end{equation}
Where $pred_{j}^r$ represents the predicted classification score of positive lane anchor $j$ in the $r$-th decoder layer. We use a feed forward network to obtain $pred_{j}^r$ from output of the $r$-th decoder layer.
$S$ represents the collection of positive lane anchors for one-to-many assignment.
O2SFormer filters low quality predictions by setting predicted classification score threshold, which causes that predictions with 
high LIOU and low predicted classification score are filtered and predictions with low LIOU 
and high predicted classification score are retained. Hence, we use LIOU to weight the soft label following~\cite{zhang2021varifocalnet}. 
The classification loss is formulated as
\vspace{-0.15cm}
 \begin{equation}\label{eqn-10} 
   {Loss}_{cls}^r=\sum_{j\in S}{FL(pred_j^r,d_j^r\times LIOU_j^r)+\sum_{j\in B}{FL(pred_j^r,0)}}  
 \end{equation}
 \vspace{-0.5cm}
 \begin{equation}\label{eqn-11} 
   {Loss}_{cls}=\sum_{r=1}^{N-1}{Loss}_{cls}^r+FL(pred_{fully}^N,LIOU_{fully}^N)+\sum_{j\neq fully}{FL(pred_j^N,0)}
 \end{equation}
 Where $FL$ is Focal loss~\cite{lin2017focal}. $LIOU_{j}^r$ denote the LIOU of positive lane anchor $j$ in the $r$-th decoder layer.  
 $pred_{fully}^N$ and $LIOU_{fully}^N$ are predicted classification score and LIOU of fully positive lane anchor in the last decoder layer. $B$ is the set of negative lane anchors. Regression loss of O2SFormer is
 \vspace{-0.15cm}
 \begin{equation}\label{eqn-12} 
   \small
   {Loss}_{reg}^r=\sum_{j\notin B} \{{\lambda_{iou}\times {L}_{iou}(b_j^r,b_{gt})+\lambda_{l1}\times [{L}_{l1}((x_j^r,y_j^r),(x_{gt},y_{gt}))+{L}_{l1}(\theta_j^r,\theta_{gt})+{L}_{l1}(l_j^r,l_{gt})]}\}
   \vspace{-0.32cm}
 \end{equation}
 \vspace{-0.2cm}
 \begin{equation}\label{eqn-13} 
   {Loss}_{reg}=\sum_{r=1}^{N}{Loss}_{reg}^r  
 \end{equation}
 $b_j^r$, $(x_j^r,y_j^r)$, $\theta_j^r$ and $l_j^r$ are predicted location of $z$ sampling points, start point coordinate, 
 angle and length of positive lane anchor $j$ in the $r$-th decoder layer. $b_{gt}$, $(x_{gt},y_{gt})$, $\theta_{gt}$ and $l_{gt}$ denote the ground truth of $b_j^r$, $(x_j^r,y_j^r)$, $\theta_j^r$ and $l_j^r$ respectively. 
 $L_{iou}$ is Line-IOU loss~\cite{zheng2022clrnet} and $L_{l1}$ is smooth-$l$1 loss~\cite{girshick2015fast}. $\lambda_{iou}$ and $\lambda_{l1}$ are the weight of $L_{iou}$ and $L_{l1}$.
 Besides, we adopt segmentation loss as an auxiliary loss following~\cite{zheng2022clrnet}. The total loss of O2SFormer is
 \vspace{-0.2cm}
 \begin{equation}\label{eqn-14} 
   Loss=\lambda_{cls}\times{Loss}_{cls}+{Loss}_{reg}+\lambda_{seg}\times{Loss}_{seg} 
   \vspace{-0.15cm}
 \end{equation}
 Where $\lambda_{cls}$ and $\lambda_{seg}$ are the weight of classification loss and segmentation loss respectively. 
 We take the Cross entropy loss as the segmentation loss.
 \vspace{-0.35cm}
 \section{Experiment}
 \subsection{Experimental Setting}
 {\bf Dataset}. We evaluate O2SFormer on the CULane dataset~\cite{pan2018spatial}. CULane is a widely used large-scale dataset for lane detection. 
 It contains a lot of challenging scenarios such as crowded roads. The CULane dataset 
 consists of 88.9K images for training, 9.7K images in the validation set, and 34.7K images for the test. Image size is 1640×590. 
 
 {\bf Evalutaion Metrics}. We adopt the F1 score to measure the performance on CULane: $F_1=\frac{2\times P r e c i s i o n\times R e c a l l}{Precision+Recall}$, where $Precision=\frac{TP}{TP+FP}\ $ and $Recall=\frac{TP}{TP+FN}$. 
 $FP$, $TP$ and $FN$ are false positives, true positives and false negatives respectively.
 \vspace{-0.3cm}
 \subsection{Implementation Details}
 We adopt ResNet~\cite{he2016deep} pretrained on ImageNet~\cite{deng2009imagenet} as the backbone. 
 All images are resized to 320×800. For data augmentation, we use random horizontal 
 flips and random affine transforms(translation, rotation and scaling). We utilize 
 AdamW~\cite{loshchilov2017decoupled} optimizer with learning rate of 2.5e-4. Cosine decay strategy is applied to 
 update learning rate. We train 20 epochs on CULane. All experiments are conducted on 
 2 GPUs. O2SFormer is implemented based on MMDetection~\cite{mmdetection}. We 
 set the number of sampling points $z$ to 72 and the number of lane anchors to 192. 
 $\lambda_{cls}$, $\lambda_{iou}$, $\lambda_{l1}$ and $\lambda_{seg}\ $ are set to 2, 2, 0.3 and 1 respectively.
 The initial value of $z$ offsets are set to 0. We initialize the position of lane anchors following~\cite{zheng2022clrnet}.
 All feed forward networks in O2SFormer consist of three MLP modules.
 \vspace{-0.2cm}
 \subsection{Comparison with Existing methods}
 {\bf Main results}. The performance of O2SFormer on CULane testing set is shown in Table~\ref{sec:table}. O2SFormer 
  achieves the best results among Transformer-based detectors. For example, O2SFormer(ResNet18) 
  outperforms the Laneformer(ResNet18) in total F1 score(76.07\% vs 71.71\%). With ResNet18, O2SFormer surpasses LSTR by a great margin (76.07\% vs 64.00\%). Compared with CNN-based detectors, O2SFormer with ResNet50 is 
  superior to LaneATT with a large ResNet122 backbone in terms of F1 score and MACs(77.83\% vs 77.02\%, 27.51G vs 86.5G). 
  Moreover, O2SFormer achieves the better performance in hard scenarios like Dazzle and Night. Visual results can be found in the supplementary materials.
 
\begin{figure*}[t]
    \begin{center}
    \includegraphics[height = 7cm,width=9cm]{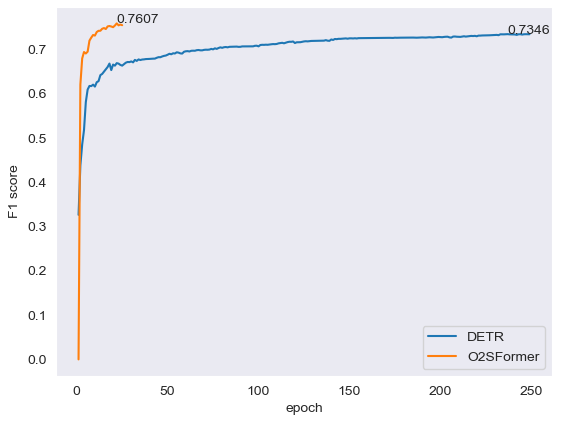}
    \vspace{-0.5cm} 
    \end{center}
       \caption{Convergence curves for O2SFormer and DETR. Two models are trained with ResNet18. O2SFormer converges much faster than DETR.}
       \vspace{-0.1cm}
    \label{fig:short4}
    \end{figure*}

 {\bf Analysis on convergence speed}. We show the comparison of convergence speed between O2SFormer and DETR in Fig.~\ref{fig:short4}. 
  Fig.~\ref{fig:short4} shows that O2SFormer converges 12.5× faster than DETR.
 \begin{table}[!h]
    \centering
    \caption{Comparison of F1 score and MACs(multiply–accumulate operations) on CULane testing set. * represents O2SFormer with hybrid-encoder used in RT-DETR~\cite{lv2023detrs}. We report false positives for “Cross” category.}
    \tabcolsep=0.5cm
    \Huge
    \resizebox{\textwidth}{!}{
    \begin{tabular}{cccccccccccc}
      \toprule
      Methods& Normal& Crowed& Dazzle& Shadow& No line& Arrow& Curve& Night& Cross& Total& MACs(G) \\
      \midrule
      LSTR(ResNet18)~\cite{liu2021end}& -& -& -& -& -& -& -& -& -& 64.00& - \\
      SCNN(ResNet50)~\cite{pan2018spatial}& 90.60& 69.70& 58.50& 66.90& 43.40& 84.10& 64.40& 66.10& 1900& 71.60& - \\
      UFLD(ResNet34)~\cite{qin2020ultra}& 90.70& 70.20& 59.50& 69.30& 44.40& 85.70& 69.50& 66.70& 2037& 72.30& - \\
      PINet(Hourglass)~\cite{ko2021key}& 90.30& 72.30& 66.30& 68.40& 49.80& 83.70& 65.20& 67.70& 1427& 74.40& - \\
      CurveLane-L~\cite{xu2020curvelane}& 90.70& 72.30& 67.70& 70.10& 49.40& 85.80& 68.40& 68.90& 1746& 74.08& 86.5 \\
      LaneATT(ResNet122)~\cite{tabelini2021keep}& 91.74& 76.16& 69.47& 76.31& 50.46& 86.29& 64.05& 70.81& 1264& 77.02& 70.5 \\
      Laneformer(ResNet18)~\cite{han2022laneformer}& 88.60& 69.02& 64.07& 65.02& 45.00& 81.55& 60.46& 64.76& 25& 71.71& 13.8 \\
      Laneformer(ResNet50)~\cite{han2022laneformer}& 91.77& 75.74& 70.17& 75.75& 48.73& 87.65& 66.33& 71.04& {\bf 19}& 77.06& 26.2 \\
      \midrule
      O2SFormer(ResNet18)& 91.89& 73.86& 70.40& 74.84& 49.83& 86.08& 68.68& 70.74& 2361& 76.07& 15.21 \\
      O2SFormer(ResNet34)& 92.50& 75.25& 70.93& 77.72& 50.97& 87.63& 68.10& 72.88& 2749& 77.03& 25.05 \\
      O2SFormer(ResNet50)& 93.09& 76.57& 72.25& 76.56& 52.80& 89.50& 69.60& 73.85& 3118& 77.83& 27.51 \\
      O2SFormer(ResNet50)*& {\bf 93.89}& {\bf 82.03}& {\bf 80.29}& {\bf 83.89}& {\bf 71.46}& {\bf 92.11}& {\bf 76.21}& {\bf 80.48}& 3208& {\bf 78.00}& 43.11 \\
      \bottomrule
    \end{tabular}}
    \label{sec:table}
  \end{table}
 
  \subsection{Ablation study}
  In this section, we ablate the key components in O2SFormer. 
  If not specific, we adopt O2SFormer with ResNet18 for all ablation experiments and results are shown in Table~\ref{sec:table2}. 
  More ablation studies can be found in supplementary materials.
  \begin{table}[htbp!]
     \centering
     \caption{Ablation study results of key components of O2SFormer on CULane. FPS is test on a single 2080Ti GPU with TensorRT.}
     \tabcolsep=0.5cm
     \huge
     \resizebox{\textwidth}{!}{
     \begin{tabular}{lccccccc}
       \toprule
       Methods& Epoch& F1(\%)& Precision(\%)& Recall(\%)& FPS& Params(M) \\
       \midrule
       DETR& 250& 73.46& 79.26& 61.79& 89& 30.9 \\
       \midrule
       +Dynamic anchor-based positional query& 50& 74.09& 81.11& 62.28& 84& 31.1 \\
       +One-to-Several label assignment& 20& 75.20& 82.88& 65.39& 84& 31.1 \\
       +Layer-wise soft label& 20& 76.07& 83.92& 65.94& 84& 31.1 \\
       \bottomrule
     \end{tabular}}
    \label{sec:table2}
   \end{table}
  \begin{figure*}[htbp!]
     \begin{center}
     \includegraphics[height = 4cm,width=6cm]{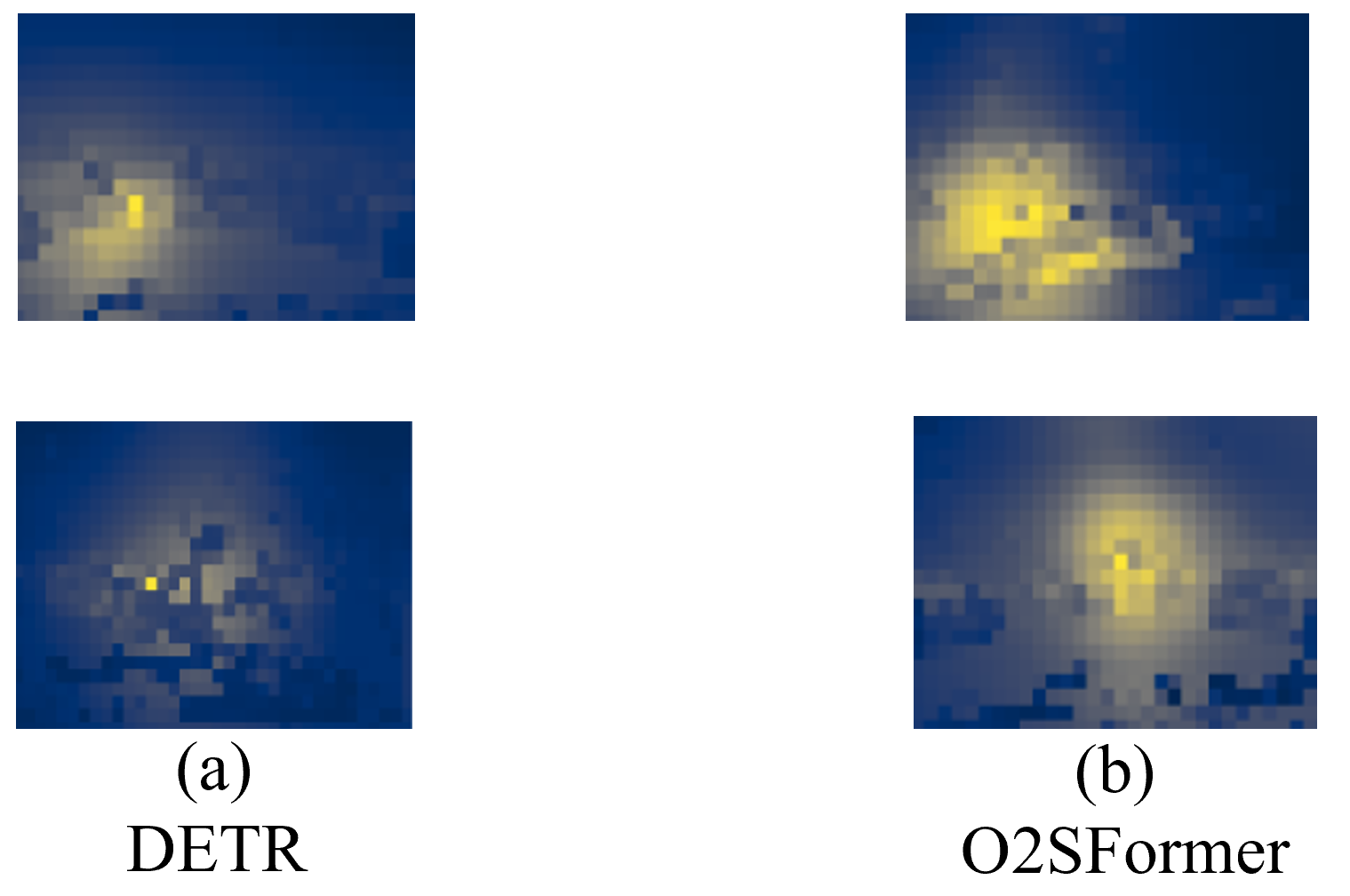}
     \vspace{-0.5cm} 
     \end{center}
        \caption{We visual the positional attention maps of positional queries for DETR and O2SFormer with similar query position. 
                 Following~\cite{liu2022dab}, we perform dot product between positional embeddings from a feature map and a positional 
                 query to calculate the attention map.}
        \vspace{-0.3cm}
     \label{fig:short5}
     \end{figure*}
  
  {\bf Effectiveness of Dynamic anchor-based positional query}. As we can see in Table~\ref{sec:table2}, 
  directly applying DETR in lane detection only obtains 73.46\% F1 score. The add of 
  Dynamic anchor-based positional query improves the F1 score from 73.46\% to 74.09\%. 
  The precision and recall are increased by 1.85\%(81.11\% vs 79.26\%) and 0.49\%(62.28\% vs 61.79\%). We also visual few 
  positional attention maps of positional queries in Fig.~\ref{fig:short5}. Compared with 
  positional attention map of DETR, the positional attention map of O2SFormer has a more clear focus area. 
  Above results demonstrate that Dynamic anchor-based positional query can effectively improve the model performance.
  
  {\bf Effectiveness of One-to-several label assignment}. One-to-one label assignment in DETR results in label semantic 
   conflicts while one-to-many label assignment needs NMS to remove duplicate predictions. We propose 
   one-to-several assignment to combine the advantages of both label assignments. As shown in Table~\ref{sec:table2}, 
   equipped with one-to-several assignment, F1 score gains 1.11\%(75.20\% vs 74.09\%) improvements 
   and recall is increased by 3.11\%(65.39\% vs 62.28\%). 
   It should be noted that one-to-several label assignment spends fewer training epochs achieving better performance.
  
  {\bf Effectiveness of layer-wise soft label}. We futher explore the impact of layer-wise soft label. 
  Layer-wise soft label delivers a 0.87\% F1 score improvements(76.07\% vs 75.20\%) and increases 
  the precision by 1.04\%(83.92\% vs 82.88\%). Since layer-wise soft label is used in training, it has no influence on FPS and parameters.

 \section{Conclusion}
 In this paper, we present the One-to-Several Transformer(O2SFormer) to achieve end-to-end lane detection. 
 Through one-to-several label assignment, we enable the detectors to address label semantic conflicts and keep end-to-end detection at the same time.
 We assign layer-wise soft label to positive lane anchors to achieve the balance between one-to-one and one-to-many assignment. 
 To explore explicit positional prior information, we propose the dynamic anchor-based positional query by encoding lane anchors as positional queries.
 Experimental results show that O2SFormer significantly speeds up the convergence of DETR and surpasses Transformer-based and CNN-based
 detectors on the CULane dataset. In the future, we will focus on applying one-to-several label assignment to CNN-based detectors.
\clearpage
\bibliography{egbib}
\clearpage
\appendix
\section*{Appendix}

In this appendix, we provide more training details(Appendix A), 
ablation studies(Appendix B) and visualization results(Appendix C) on CULane dataset.

\renewcommand{\thesubsection}{\Alph{subsection}}
\subsection{More training details}

Following~\cite{zheng2022clrnet}, the cost of OTA includes classification cost $C_{cls}$ 
and similarity cost $C_{sim}$. The total cost is defined as:
\begin{equation}\label{eqn-15} 
   Cost=w_{sim}\times C_{sim}+w_{cls}\times C_{cls}
   \vspace{-0.5cm}
 \end{equation}
 \begin{equation}\label{eqn-16} 
   C_{sim}={(C_{dis}\times C_{xy}\times C_{theta})}^2
 \end{equation}
 Here we take the focal cost~\cite{lin2017focal} between predictions and labels as $C_{cls}$. $C_{sim}$ contains three components, $C_{dis}$ 
 denotes the average distance between valid sampling points and ground truth, $C_{xy}$\ means the distance of the 
 start point between predictions and ground  truth, $C_{theta}$\ means the difference of the angle between predictions 
 and ground truth. We normalize the $C_{theta}$\ to [0,1]. $w_{cls}$ And $w_{sim}$ are the weight of classification cost 
 and similarity cost. We set $w_{sim}$ and $w_{cls}$ to 3 and 1 respectively. Besides, we set the number of decoder layer N to 6.

\subsection{More ablation studies}

In this section, we provide more ablation study results to confirm the effectiveness of O2SFormer. 
If not specific, we still adopt the ResNet18 as the backbone.
\vspace{-0.3cm}
\begin{table}[!h]
   \centering
   \caption{Results of number of lane anchors.}
   \tabcolsep=0.5cm
   \small
   \resizebox{0.67\textwidth}{!}{
      \begin{tabular}{ccc}
         \toprule
         Number of lane anchors& F1(\%)& MACs(G) \\
         \midrule
         4& 68.25& 11.92 \\
         16& 71.48& 12.13 \\
         64& 73.69& 12.97 \\
         192& 76.07& 15.21 \\
         256& 76.10& 16.34 \\
         512& 76.28& 20.82 \\
     \bottomrule
     \vspace{-0.4cm}
   \end{tabular}}
 \end{table}

 {\bf Number of lane anchors}. We ablate the number of lane anchors and results are shown in Table 3. Since there are at most 4 lanes in a image, 
 we increase the number of lane anchors from 4 to 512. We can obviously observe that as the number of lane anchors increases, 
 F1 score improves significantly. When number of lane anchors is 256 or 512, MACs is increased from 15.21G to 
 16.34G or 20.82G, whereas F1 score improves slightly. Hence, we set the number of lane anchors to 192.

 \begin{table}[!h]
   \centering
   \caption{Performances of different one-to-many label assignments.}
   \tabcolsep=0.5cm
   \small
   \resizebox{0.67\textwidth}{!}{
      \begin{tabular}{cccc}
         \toprule
         Method& F1(\%)& Precision(\%)& Recall(\%) \\
         \midrule
         ATSS~\cite{zhang2020bridging}& 74.21& 81.59& 64.77 \\
         PAA~\cite{kim2020probabilistic}& 75.19& 82.76& 65.98 \\
         OTA~\cite{ge2021ota}& 76.07& 83.92& 65.94 \\
     \bottomrule
     \vspace{-0.4cm}
   \end{tabular}}
 \end{table}
 {\bf Performances of different one-to-many label assignments}. Table 4 compares different one-to-many label assignments. We can see that OTA achieves better performance 
 than the other two label assignments in terms of F1 score and Precision. 
 Therefore, we adopt the OTA as the one-to-many label assignment in O2SFormer.

\subsection{Visualization results on CULane}
In Fig.6, we visualize the detection results of O2SFormer on CULane. 
O2SFormer can effectively detect lanes under various scenarios.
\begin{figure*}[!h]
   \begin{center}
   \includegraphics[height = 10cm,width=11cm]{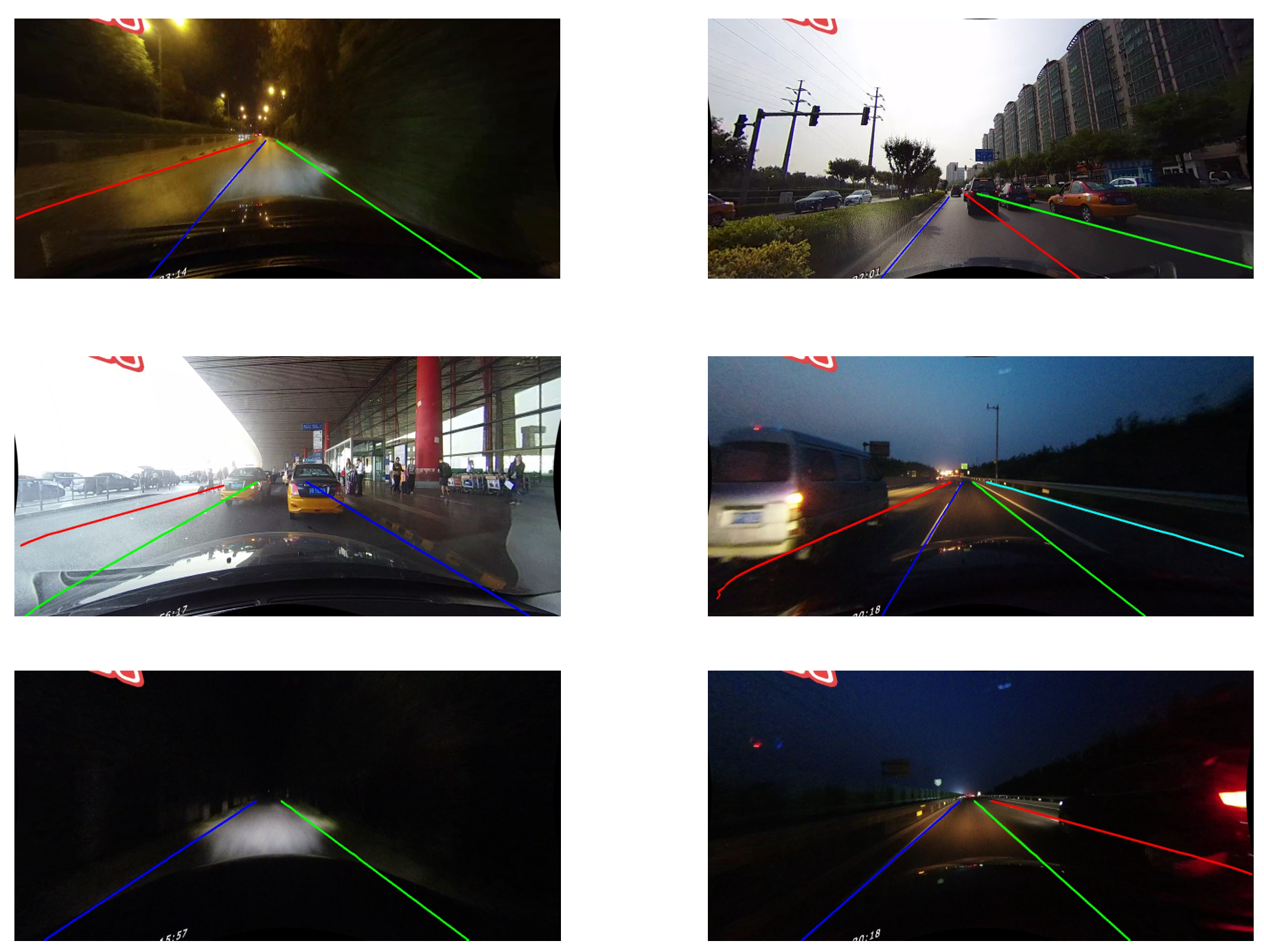}
   \vspace{-0.5cm} 
   \end{center}
      \caption{Viusalization results on CULane.}
      \vspace{-0.1cm}
   \label{fig:short}
   \end{figure*}

\end{document}